\documentclass[runningheads]{llncs}
\usepackage[T1]{fontenc}
\usepackage{graphicx}
\begin{document}
\title{Knowledge Discovery using Unsupervised Cognition}
%
%
\author{Alfredo Ibias\inst{1}\orcidID{0000-0002-3122-4272} \and
Hector Antona\inst{1} \and
Guillem Ramirez-Miranda\inst{1}\orcidID{0000-0003-2741-3705} \and
Enric Guinovart\inst{1}
}
\authorrunning{A. Ibias et al.}
%
\institute{Avatar Cognition, Barcelona, Spain\\
\email{\{alfredo, hector, guillem, enric\}@avatarcognition.com}
}
\maketitle              
\begin{abstract}
Knowledge discovery is key to understand and interpret a dataset, as well as to find the underlying relationships between its components. Unsupervised Cognition is a novel unsupervised learning algorithm that focus on modelling the learned data. This paper presents three techniques to perform knowledge discovery over an already trained Unsupervised Cognition model. Specifically, we present a technique for pattern mining, a technique for feature selection based on the previous pattern mining technique, and a technique for dimensionality reduction based on the previous feature selection technique. The final goal is to distinguish between relevant and irrelevant features and use them to build a model from which to extract meaningful patterns. We evaluated our proposals with empirical experiments and found that they overcome the state-of-the-art in knowledge discovery.

\keywords{Knowledge Discovery \and Pattern Mining \and Dimensionality Reduction \and Feature Selection \and Unsupervised Learning.}
\end{abstract}
\section{Introduction}
Knowledge discovery is a fundamental task in a handful of fields, from data science to knowledge engineering. Discovering knowledge is a difficult task that consist in extracting, from the source data, new knowledge not previously known to the practitioner. This knowledge can come in multiple forms and can be discovered through different ways. In this paper we will focus on two already established frameworks for discovering knowledge: pattern mining and feature selection. This knowledge can have multiple uses too, and in this paper we will focus in its use for reducing the dimensionality of the data. Finally, we will approach all these problems using as base the Unsupervised Cognition algorithm~\cite{iarga24}, that is a novel pattern matching, unsupervised learning algorithm that focuses on modelling the associations between the different features of the data. These characteristics allow it to build internal representations that contain the associativeness of data features, which makes them useful to discover association rules.

Pattern mining is the knowledge discovering framework that encompasses all techniques that aim to discover association rules that describe the original data. The end goal of this framework is to be able to find patterns that help practitioners to understand and interpret the original data, as well as to help them in tasks as diverse as decision-making, identification, classification, clustering or prediction~\cite{fgwnstd22}. The pattern mining techniques can be organised into two types: static and incremental. The static pattern mining techniques take the whole dataset and process it to find the patterns, and thus anytime the dataset is updated the whole process has to be run from the start to find the updated patterns. The incremental pattern mining techniques build the patterns somewhat incrementally, and thus anytime the dataset is updated they execute a set of operations that update the patterns accordingly. In this paper we present an incremental pattern mining technique. Our proposal consist in selecting a subset of the representations of a trained Unsupervised Cognition model as the patterns of the data. This subset should represent the whole dataset, and at the same time each sample from the dataset should be identified by only one representation of such subset.

Feature selection is the knowledge discovering framework that encompasses all techniques that aim to reduce the amount of features of a dataset by keeping the most relevant ones for the task at hand. The end goal of this framework is to be able to provide a list of data features that will help the practitioner to solve the task at hand, while discarding the features that only hampers the efforts to solve it~\cite{ildmpw22}. For example, a feature that clearly identifies samples of one class in a classification problem would be a selected feature, while a feature that introduces noise and does not help to identify samples of any class in that same classification problem would be a discarded feature. In this paper we present a feature selection technique that is built over our pattern mining technique. Our proposal consist in selecting features based on their correlation with the target feature, calculated over the patterns mined from a trained Unsupervised Cognition model.

Dimensionality reduction is a critical technique when dealing with huge datasets with many features. The end goal is to be able to better discriminate between the classes present in the data, avoiding noise and distilling the relevant information for the task at hand~\cite{aht20}. The dimensionality reduction algorithms can be of two types: those that transform the data and those that only select subsets of the data. The algorithms that transform the data reduce dimensionality by projecting the data into a new space with lower dimensionality, thus loosing some information, but taking into account all the collected features. The algorithms that select subsets of the data reduce dimensionality by discarding features, thus loosing the information of those discarded features, but keeping all the information of the remaining ones. In this paper we present a dimensionality reduction algorithm of the second kind, using our proposal of feature selection for deciding which features are the ones that will remain after the reduction of dimensionality. Our proposal consist in taking the features selected with our feature selection proposal and reduce the dataset to have only those features.

Presenting these proposals, the aim of this paper is to show a viable pipeline to use the result of a trained Unsupervised Cognition model for knowledge discovery. The final purpose is to help the practitioner to extract as much knowledge as possible from the use of the Unsupervised Cognition algorithm, taking advantage of its particular characteristics. This knowledge can be later used by the practitioner to better understand the data and the underlying relationships it comprises, which can be useful in areas as diverse as medicine~\cite{hld12,khdzlc23,rdp22,tapt21}, business analytics~\cite{dm06}, or user profiling~\cite{aglmpp12}. To that end, using our proposals we have elaborated a knowledge discovery pipeline that consists in using the feature selection technique to select the truly relevant features of the dataset, then use these features to perform dimensionality reduction to train a new Unsupervised Cognition model, and finally use that newly trained Unsupervised Cognition model to extract relevant patterns, with our pattern mining proposal, for the practitioner to explore.

To evaluate the performance and usefulness of our proposals, we have performed multiple experiments. The aim of those experiments is to validate our proposals and show how to use them in a practical scenario. Specifically, we performed three experiments, one per proposal. The pattern mining experiment evaluates the validity of the produced patterns, the dimensionality reduction experiment evaluates the effectiveness of our dimensionality reduction proposal, and the feature selection experiment evaluates the validity of the selected subset of features. The three proposals are evaluated using Unsupervised Cognition models for consistency, and the second one is compared with state-of-the-art methods. The results of the experiments show that our proposals are better than current state-of-the-art, and that their results are useful for the practitioner. Moreover, they show that using our proposed pipeline, practitioners can increase the accuracy of their Unsupervised Cognition models.

The rest of the paper is organised as follows: First, in Section~\ref{sec:relwork} we present the work related with our proposals, to be followed by our pattern mining proposal in Section~\ref{sec:pattern}, our dimensionality reduction proposal in Section~\ref{sec:dimension}, our feature selection proposal in Section~\ref{sec:feature}, and our knowledge discovery pipeline proposal in Section~\ref{sec:pipeline}.
Then, in Section~\ref{sec:exp} we describe our experiments and their results, including a discussion about the threats to their validity in Section~\ref{sec:ttv}.
And finally, in Section~\ref{sec:conc} we present the conclusions of our work.

\section{Related Work}\label{sec:relwork}
In the pattern mining field there are multiple techniques~\cite{fgwnstd22}, from graph-based pattern mining techniques~\cite{fhclzly20} to heuristic pattern mining ones~\cite{dc17,szhl22}. However, most of these techniques should be applied over the original data, or are built over specific methods. In this paper we are presenting a novel pattern mining technique specifically defined for an Unsupervised Cognition model, thus not being comparable to the others in its conception. Moreover, comparing patterns is already a hard task, as they usually represent different associations between the features of the original data. Finally, our patterns are of the form of the original data, while most pattern mining algorithms return association rules. Thus, we are not aware of any pattern mining technique able to extract patterns from an Unsupervised Cognition model and that present those patterns as elements of the dataset.

In the feature selection field there are multiple techniques~\cite{ildmpw22}, but most of them follow the same structure: they start removing features based on which feature improves more a certain measure when removed. Thus, they mostly work by trial and error, with some methods to avoid the combinatorial explosion problems. In contrast, our method is one of the few that are based on correlations or similar measures to order the features and select a subset of them based on a threshold. Moreover, our method is developed exclusively for its use over the patterns extracted from an Unsupervised Cognition model, and thus it has no comparison with other methods, as up to date there are no other proposals for feature selection over the results of an Unsupervised Cognition model.

In the dimensionality reduction field there are multiple techniques~\cite{aht20}, mainly divided into two groups: linear and non-linear. The linear techniques use simple, linear transforms to reduce the dimensionality of the data, while non-linear techniques use non-linear transforms. In the case of linear techniques, the most well known and tested technique is PCA~\cite{pear01}, a traditional method that is still the state-of-the-art on the field~\cite{aht20}. In the case of non-linear techniques, the current state-of-the-art is an unsupervised clustering algorithm called Self Organising Maps (SOM)~\cite{kohonen98}. Both algorithms transform the original data into a two-dimensional space, thus losing a lot of information in the way. However, they are the current state-of-the-art for dimensionality reduction and thus we will compare our proposal versus them.

\section{Pattern Mining}\label{sec:pattern}
Unsupervised Cognition~\cite{iarga24} is an unsupervised learning method that builds representations in a hierarchical structure. To briefly resume its functioning, it builds representations by combining multiple inputs, thus each representation depicting a subset of the input domain. Then, for each representation, it builds children representations that depict subsets of the set depicted by the parent representation, until there is no more possible subsets. Moreover, each input is assigned to a representation based on a similarity measure, thus each representation depicts a cluster of the input domain.
This way, the Unsupervised Cognition algorithm builds a tree of representations where the seed \emph{Cell} contains the most generic representations, that is, those that depict the biggest clusters of the input domain; and the leaf Cells contain representations that depict only one input, that is, literal representations. Here it is important to remark that representations at the same level of tree depth do not share any input, that is, any input will belong to only one representation at that level.

These representations are built combining multiple inputs through average sum. To know more details about how these representations are built (i.e. details about how the algorithm decides to which representation a new input belongs) we refer to the original paper~\cite{iarga24}. We also refer to it~\cite{iarga24} to assess the validity of the representations built by the Unsupervised Cognition algorithm. There their validity is shown through their effectiveness as part of a pattern matching algorithm. For this paper we only need to know that the resulting structure of the training of an Unsupervised Cognition model is a tree of representations where the parent-children relationship is that of a set and its subsets, and the difference between representations at the same level is based on their similarity.

Our proposal is to extract patterns from the representations built by the Unsupervised Cognition algorithm. To do so, first we need to define what we consider to be a pattern. In our case, we will consider patterns those representations that are the most generic ones. Under no constraints, the most generic representations build by the Unsupervised Cognition algorithm are those present in its seed Cell, as they are the ones that aggregated the most quantity of inputs. At the same time, these representations have split the input domain into multiple clusters (one per representation) based on similarity distance, thus telling us that there are different forms an input can take. And due to the fact that representations at the same tree depth do not share inputs, then these representations are the different patterns present in the data.

However, if we have some constraints, then the subset of representations that we will consider patterns change. For example, if we have trained an Unsupervised Cognition model with the purpose of solving a classification task, then the representations will include information of the class they belong (even though the learning algorithm is fully unsupervised). In this case, it is common that some representations at multiple levels of the tree belong to more than one class, because they combined inputs belonging to different classes. Thanks to the set - subset relationship of the parent and children representations, we know that those representations that belong to multiple classes are in the top of the tree, while the representations that belong to only one class are at the bottom of the tree. In this situation, we may want to add the constraint that the patterns differentiate between classes. Given the previous organisation, in this case we will consider patterns those representations that are in the limit, that is, those that belong to only one class, but that at the same time are the most generic ones possible. These representations will have parent representations that belong to multiple classes (unless they are in the seed Cell) and all their children representations (if they have any) will belong to the same class than them. This setup provides us with at least one pattern for each class, and we may have multiple patterns for the same class.

A relevant remark about these patterns is that they are an average sum of the inputs, and thus they are highly interpretable. Thanks to this, they can help to understand the relationship between the input features and their associations. Moreover, they are a structured view of the input domain, where the different patterns split the input domain into separated subdomains. This would allow the practitioner to better understand the typology of the domain. If we factor in the fact that the learning algorithm is unsupervised, it is clear that the resulting patterns will not be influenced by possible class labels, and thus we may have multiple patterns for the same class that represent the different forms in which the elements of such class can appear. This is useful to fully grasp the knowledge about a class, as knowing its different possible forms helps to understand all the ways a class can appear.

Finally, we want to remark that the fact that the representations are based on similarity distance implies that the found patterns are substantially different from one another. This in fact includes special patterns for outliers, which allow us to recognise them as a separate pattern from the rest of the data. In the end, we end with few patterns that represent the whole input domain, with each input belonging to only one pattern. This will be useful to understand the input domain, as well as to avoid biases based on the repetition of the data in the learning dataset, allowing the practitioner to focus only on the data characteristics. We would also like to remark that this approach is a stream pattern mining algorithm, due to its use of the internal representations. Whenever a new input is provided to the algorithm, it updates its representations, thus updating its patterns.

\section{Dimensionality Reduction}\label{sec:dimension}
Once we have a set of patterns selected with respect to a target feature, we propose to use them to reduce the dimensionality of our input domain. This would be useful in situations where we have plenty of features for each input, but not all are relevant for the task at hand. For example, if we have recollected a lot of genetic data, and we want to solve a classification problem where only a handful of genes are involved, it is very probable that not only most of the genes collected in the genetic data will not help to solve that task, but that they will actually hurt the efforts to solve the problem adding unnecessary and irrelevant noise. Thus, reducing data dimensionality is a fundamental step in these kind of cases.

Ideally, the goal is to reduce the dimensionality of the input domain by removing the features from the inputs that are less relevant for the task at hand. To that end, first we need to have a target feature, that for the Unsupervised Cognition algorithm should be a metadata feature. This is a prerequisite because our dimensionality reduction method will be based in how much a feature correlates with such target feature. The rationale behind our dimensionality reduction method is that features that correlate with the target feature may contain relevant information to solve the task at hand, while features that do not correlate with the target feature are irrelevant noise.

With this rationale, we devised the following method: we start by picking the patterns generated by the Unsupervised Cognition algorithm, and we compute, for each feature, the correlation between the values of such feature in the patterns and the value of the target feature in the patterns. If the target feature is a numerical feature, the patterns will be the representations from the seed Cell of the Unsupervised Cognition model. However, if the target feature is a categorical feature, then the patterns should be those representations that belong to only one class of the classes present in the categorical feature. In this last case, the correlations will be between each class of the target feature and the values in the other feature, thus generating a correlation for each class of the target feature, for each feature of the input domain.

With the correlations, and their corresponding p-values, computed, we filter out those features whose correlation has a p-value higher than $0.05$, as it is usually done in statistics. Then, between the remaining correlations, we filter out those that are lower than a certain threshold. This threshold will depend on the obtained correlations, and it should be set with the goal of selecting few features, but enough features to be able to solve the task at hand. In any case, this threshold should be higher than an absolute value correlation of $0.5$. If no correlation is higher than $0.5$, then we have no relevant features for the given target feature, and thus we cannot reduce the dimensionality in a meaningful manner.

Finally, once we have few features left as the relevant features, we have our final list of features selected for reducing the dimensions of the input domain. Thus, we can use them to train a new Unsupervised Cognition model only with the data of those relevant features, and produce new, simpler but more meaningful, patterns. Moreover, we expect that selecting these relevant features, and filtering out those that are not relevant, the accuracy obtained by Unsupervised Cognition will increase, as it will not be biased by irrelevant noise.

As you can see, the dimensionality reduction method that we are proposing is nothing new. Moreover, it has been used plenty of times to reduce the dimensionality of data, and it is considered good practice in the Machine Learning community~\cite{pear01}. However, the difference of our proposal lies in the fact that we are not computing correlations over the raw data. Instead, we are computing correlations over the patterns obtained from an Unsupervised Cognition model, which aims to build representations that model the data in the learning dataset. Thus, we have fewer points to compute the correlations but at the same time the points we have are more meaningful, as they are from patterns that represent a subset of the input domain. Moreover, this avoids the bias that can be induced in the correlations when there are more samples of one pattern than of the others.

This bias is important because when computing the correlations with the raw data, having more samples of one pattern than of the others will force the correlation to have more into account such pattern due to its size. In our case, as each pattern has only one point for computing the correlations, this bias is not present, thus computing the correlations as if all the patterns were equally probable. We consider this last behaviour preferable in most situations, where the limitations of data recollection may influence the results of how many samples of each pattern appears in the learning dataset. This is specially critical for small datasets, where it is more probable that the recollected data does not fully represent the real-world population distribution; and in cases where the population distribution is not relevant due to the nature of the problem.

\section{Feature Selection}\label{sec:feature}
An important property of our dimensionality reduction proposal is that it is feature based. That means that it does not transform the data between mathematical spaces like other dimensionality reduction methods do, but instead it only selects a set of relevant features. Thus, it is in fact a feature selection method too. However, our proposal has a caveat: even though the Unsupervised Cognition algorithm is deterministic, the patterns it produces are subject to the order in which the data is provided to the algorithm. Thus, different data orders produce different patterns and thus different sets of relevant features, although there are many repeated features between them.

Given this problem, here we propose a basic solution: to select relevant features from multiple Unsupervised Cognition models, each one trained with a different input order. Ideally, we should train at least $100$ Unsupervised Cognition models, and extract a list of relevant features from each one using the mechanism presented in the previous section. Once we have a set of lists of relevant features, one for each trained Unsupervised Cognition model, we can count in how many lists appears each feature. This count normalised provides us with a \emph{confidence} measure to evaluate how probable is that each feature is relevant for the task at hand. Thus, now we can set another threshold and select as relevant only those features that are over the threshold. Here we recommend a threshold of at least $50\%$ confidence, as lower confidence means that in most of the cases that feature was not relevant for the task at hand. Ideally, the threshold would be $100\%$ confidence, meaning that such features were relevant in all cases.

A downside of this approach is that there is not always a feature that has $100\%$ confidence, remarking the need for this approach but at the same time showing the huge dependency of the Unsupervised Cognition algorithm in their learning dataset input order. This stems from the fact that such algorithms are world modelling algorithms, and thus depend on how the world is presented to them. Nonetheless, this world modelling aspect of the Unsupervised Cognition algorithm is also the main factor why we decided to develop our proposals over these algorithms, because we considered that the patterns they build are going to be more robust and representative than the patterns build by other algorithms that are not based on modelling.

\section{Knowledge Discovery Pipeline}\label{sec:pipeline}
First of all, we want to emphasise that, although we have presented our proposals in dependency order, our ideal pipeline for knowledge discovery will be to use the presented feature selection technique to select the truly relevant features of the dataset, then use these features to perform dimensionality reduction to train a new Unsupervised Cognition model, and finally use that newly trained Unsupervised Cognition model to extract relevant patterns, with our pattern mining proposal, for the practitioner to explore. We have to remark here that the feature selection technique already uses the pattern mining and dimensionality reduction techniques inside it, hence why we presented those techniques first even although in our pipeline proposal they are used in reverse order.

\begin{figure}[t]
    \centering
    \includegraphics[width=1\columnwidth]{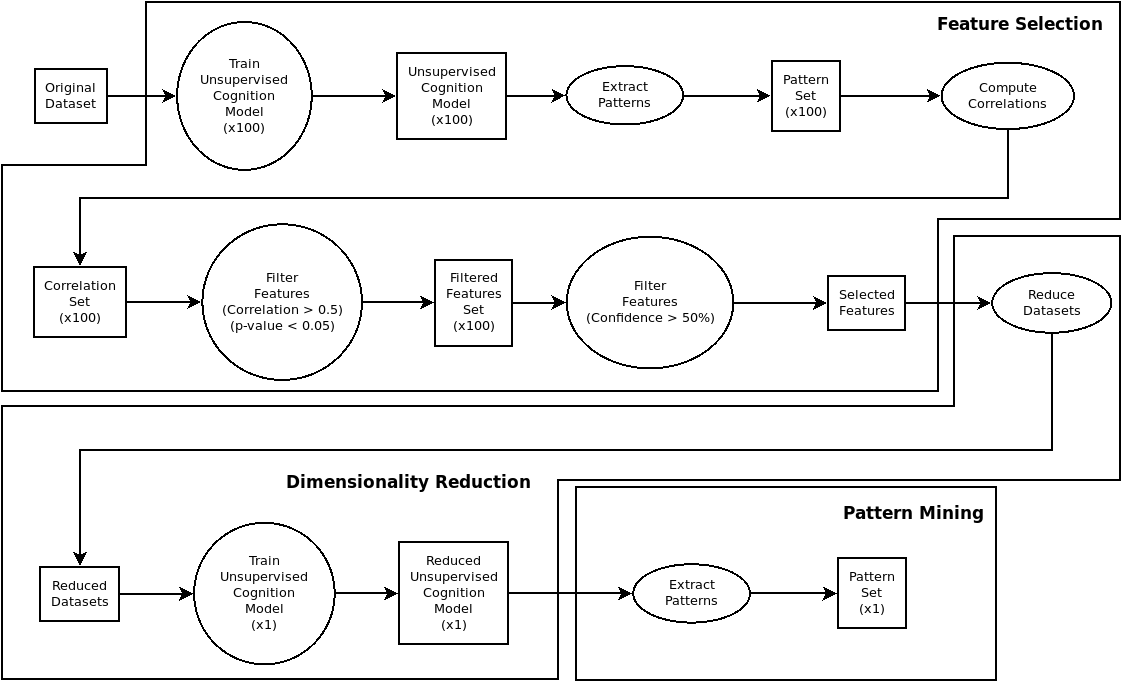}
    \caption{Pipeline outline.}
    \label{fig:pip}
\end{figure}

To be more precise, our proposed knowledge discovery pipeline comprises the following steps (an outline of the pipeline is displayed at Figure~\ref{fig:pip}):
\begin{itemize}
    \item Train a significant amount of Unsupervised Cognition models (at least $100$) over the original train dataset. Each model will receive the dataset in a different order, thus building a different internal representation. We can also test the models over the original test dataset to know the average test accuracy.
    \item Extract the set of patterns from each Unsupervised Cognition model and compute the correlations for each set.
    \item Filter, for each pattern set, the features based on their correlation and p-value, to obtain a confidence value for each feature.
    \item Filter the features based on their confidence value to select a final set of features.
    \item Reduce the train and test datasets removing the features that were not selected in the previous step.
    \item Train an Unsupervised Cognition model over the reduced train dataset. We can also test the resulting model over the reduced test dataset to know its test accuracy.
    \item Extract the set of patterns from the last Unsupervised Cognition model. Exploring these patterns will be useful to discover new knowledge.
\end{itemize}

This pipeline has been designed to provide the practitioner with the best possible patterns from the original data, with the goal that such patterns are understandable and interpretable enough so the practitioner is able to explore them and discover new knowledge. In a sense, these patterns will be cluster representatives, where any similar input will belong to their cluster. However, at the same time they are aggregations of the inputs that belong to such cluster, thus having the information of all the casuistry present in each cluster. Moreover, they are associative patterns, where the association between the different features conforming the pattern has been keep unaltered. And finally, these patterns are disjointed, thus dividing the input domain into disjoint subsets. All these properties make these patterns useful to discover new knowledge, from basic associations between features that define a class, to identification and explanation of outliers.

\section{Experiments}\label{sec:exp}
To evaluate the usefulness of our proposals, we devised the following experiment: we will evaluate how good are the selected patterns to compute correlations for dimensionality reduction by comparing the accuracy of an Unsupervised Cognition model before and after reducing the dimensionality of the input domain. In an additional experiment we will evaluate the validity of our dimensionality reduction proposal by comparing our proposal with other dimensionality reduction methods by computing the accuracy of an Unsupervised Cognition model after reducing the dimensionality with each method. Finally, in our last experiment we will evaluate the validity of our feature selection proposal by comparing the accuracy of an Unsupervised Cognition model over the original data and over the selected features. 

To perform our experiments, we selected as experimental subject the TCGA Kidney Cancers dataset~\cite{wcmsoesss13}. This dataset has $58056$ features but with only $897$ entries, what makes it the ideal dataset where to perform our experiments. The idea is that, with so many more features than entries, this dataset will have a lot of noisy or irrelevant features, that are the ones we will aim to remove. We decided to perform our experiments over only one dataset because our aim is to compare methods, thus the dataset not being so critical, and because finding open, public, and accessible datasets with so many features has proven to be a really difficult task.

\subsection{Pattern Validity}
Validating the patterns that an Unsupervised Cognition generates is not a simple task. First, because the desired properties of the patterns (i.e. that they represent all the input domain, they have no overlap, etc...) are fulfilled by definition. And second because the utility of the patterns depends on the task the practitioner needs to solve. In our case, as our proposal includes a dimensionality reduction technique, we will consider that the utility of the patterns is for reducing the dimensionality of the input domain. Thus, we need to evaluate how effective this dimensionality reduction is. To measure the effectiveness of a dimensionality reduction method, the best metric is the accuracy obtained by a classification algorithm, and thus the goal will be to improve this accuracy after reducing the dimensions of the input domain. To simplify our experiments, we will use the Unsupervised Cognition algorithm as our classification algorithm.

Given this setup, the experiment we performed consisted in: training an Unsupervised Cognition model with the original train dataset, computing the accuracy of the generated model over the original test dataset, extracting the patterns from the generated model, computing the features to remove using our proposal of dimensionality reduction, removing the selected features from the training and test datasets, training an Unsupervised Cognition model with the reduced train dataset, and computing the accuracy of the resulting model over the reduced test dataset. For the dimensionality reduction we decided to keep only those features that had a correlation higher than $0.6$ with a p-value lower than $0.01$.

As explained before, one critical aspect of the Unsupervised Cognition algorithm is that, even although it is deterministic, it is subject to variations based on the order the learning dataset is provided to them. This variability affects even to the patterns they produce. Thus, to get a reliable metric of performance, we decided to repeat this experiment $100$ times, each one with a different input order, and we provide the average values as results. This implies that each dimensionality reduction selects a different set of features to keep, based on the patterns generated during the initial training.

The results obtained by this experiment were positive: starting from an average train accuracy of $0.8473$ and test accuracy of $0.8369$, we managed to reduce an average of $96.44\%$ of the dimensions to obtain an average train accuracy of $0.8847$ and test accuracy of $0.8759$. That is, reducing from $58056$ features to an average of $2067.32$ features, we obtained an average increase in accuracy of $0.0373$ in train and of $0.0391$ in test.

\subsection{Dimensionality Reduction Effectiveness}
To measure the effectiveness of our dimensionality reduction method, we decided to compare it with other dimensionality reduction methods. Specifically, we decided to compare it with PCA and Self-Organising Maps (SOM). To compare these methods, we decided to take the result of the dimensionality reduction and provide it to the Unsupervised Cognition algorithm in order to obtain a classification accuracy. The rationale behind this decision is that the dimensionality reduction method that better resumes the data should be that one that produces the higher accuracy after applying a classification method. We decided to use the Unsupervised Cognition algorithm for this task both because it was easier to use and because as an unsupervised learning algorithm it focuses on the input data, without being able to use the label to fix any errors.

Given this setup, the experiment we performed consisted in: generating a new train and test datasets using the dimensionality reduction method, training an Unsupervised Cognition model with the reduced train dataset, and computing the accuracy of the resulting model with the reduced test dataset. For our proposal, we took the results from the experiment performed in the previous section.

Same as in the previous experiment, to get a reliable metric of performance for the Unsupervised Cognition algorithm, we decided to repeat this experiment $100$ times with it, each one with a different input order, and we provide the average values as the results of the Unsupervised Cognition algorithm. That means that, for each dimensionality reduction method, we had to train $100$ Unsupervised Cognition models in order to obtain the results displayed at Table~\ref{tab:exp2}. There, we show the results after reducing dimensionality using our proposal, using PCA to reduce to $2$, $10$ and $100$ dimensions respectively, and using SOM to reduce to $2$ and $9$ dimensions respectively. We included more dimension to reduce to in order to explore how PCA and SOM can improve their results after their not so good performance with the original reduction to $2$ dimensions. However, although we tried to increase the number of dimensions to a level similar to the number of dimensions obtained by our method, the algorithms were unable to do so due to performance limitations. Thus, we decided to show some cases just for reference.

\begin{table*}[t]
\caption{Results of the reduction of dimensionality using different methods.}\label{tab:exp2}
\begin{center}\scalebox{0.85}{$
\begin{array}{| c | c | c | c | c | c | c | c |}
\hline
\mathbf{Dimensionality} & \mathbf{Initial} & \mathbf{Initial} & \mathbf{Final} & \mathbf{Final} & \mathbf{Train} & \mathbf{Test} & \mathbf{Average}\\
\mathbf{Reduction} & \mathbf{Train} & \mathbf{Test} & \mathbf{Train} & \mathbf{Test} & \mathbf{Accuracy} & \mathbf{Accuracy} & \mathbf{Dimensionality}\\
\mathbf{Method} & \mathbf{Accuracy} & \mathbf{Accuracy} & \mathbf{Accuracy} & \mathbf{Accuracy} & \mathbf{Gain} & \mathbf{Gain} & \mathbf{Reduction}\\
\hline
\mathbf{Proposal} & 0.8473 & 0.8369 & 0.8847 & 0.8759 & 0.0373 & 0.0391 & 0.9644\\
\hline
\mathbf{PCA\ 2dim} & 0.8473 & 0.8369 & 0.6212 & 0.5863 & -0.2261 & -0.2506 & 0.99997\\
\hline
\mathbf{PCA\ 10dim} & 0.8473 & 0.8369 & 0.7524 & 0.6574 & -0.0949 & -0.1795 & 0.9998\\
\hline
\mathbf{PCA\ 100dim} & 0.8473 & 0.8369 & 0.7049 & 0.6175 & -0.1424 & -0.2194 & 0.998\\
\hline
\mathbf{SOM\ 2dim} & 0.8473 & 0.8369 & 0.6304 & 0.5884 & -0.2169 & -0.2485 & 0.99997\\
\hline
\mathbf{SOM\ 9dim} & 0.8473 & 0.8369 & 0.6251 & 0.5746 & -0.2222 & -0.2623 & 0.9998\\
\hline
\end{array}$
}
\end{center}
\end{table*}

The results show how our method performs the smaller dimensionality reduction, but at the same time it is the only one able to improve the accuracy of the results. Thus, we can conclude that ours has been the only method able to distinguish the noise from the data and remove such noise properly. Moreover, we can conclude that ours is the current state-of-the-art for reducing dimensionality using the Unsupervised Cognition algorithm.

\subsection{Feature Selection Validity}
To measure the validity of the features selected by our feature selection method we devised a comparison experiment, where we compare the accuracy obtained by an Unsupervised Cognition model before and after reducing data dimensionality using the features selected by our method. The rationale behind this experiment is that a good feature selection method would be that one that selects the features better suited to produce a good classifier, that is, those that produce a higher accuracy after applying a classification method. Same as in the previous experiment, we decided to use the Unsupervised Cognition algorithm for this task both because it was easier to use and because as an unsupervised learning algorithm it focuses only on the input data.

Given this setup, the experiment we performed consisted in: training $100$ Unsupervised Cognition models with the original train dataset (each one with a different input order), computing the accuracy of the resulting models with the original test dataset, computing a subset of features with the feature selection method, generating a new train and test datasets containing only the selected features, training an Unsupervised Cognition model with the reduced train dataset, and computing the accuracy of the resulting model with the reduced test dataset. The threshold used for feature selection was a $100\%$ confidence, that is, we took only the features that had a correlation higher than $0.6$ with p-value lower than $0.01$ in all the $100$ Unsupervised Cognition models we trained.

Same as in previous experiments, to get a reliable metric of performance for the Unsupervised Cognition algorithm, we decided to repeat the last part of the experiment $100$ times with it, each one with a different input order, and we provide the average values as the results of the Unsupervised Cognition algorithm. Here it is important to remark that, in contrast with the experiments performed in the previous two sections, in this case these last $100$ Unsupervised Cognition models were trained selecting the same set of features, that were those selected by our feature selection method.

That means that, after executing our feature selection method, we trained $100$ Unsupervised Cognition models that obtained, in average, an increase in accuracy of $0.0566$ in train and of $0.0629$ in test. To be precise, the Unsupervised Cognition models trained over the original data obtained an average train accuracy of $0.8474$ and test accuracy of $0.8369$, and after reducing the dimensionality in a $99.495\%$ (from $58056$ to $291$ features) the Unsupervised Cognition models trained over the reduced data obtained an average train accuracy of $0.9039$ and test accuracy of $0.8998$.

\subsection{Threats to Validity}\label{sec:ttv}
In this section we discuss the possible threats to the validity of our results. The first kind are the threats to internal validity, that can explain our results due to uncontrolled factors. The main threat in this category is the possibility of having a faulty code. To reduce this threat we have carefully tested each piece of code used in our experiments and developed unit tests for them, and we have relied on widely tested libraries like scikit for the PCA and SOM algorithms, and the authors implementation for the Unsupervised Cognition algorithm. Another threat in this category is the impact of randomisation in the comparison results. Every time we had this problem we performed an average and standard deviation computation over multiple runs of the experiment. We performed $100$ iterations in each case due to resource constrains, but we are sure that the variation with a higher number of iterations will not be great, as the standard deviations obtained were at most of $0.03$.

The second kind of threats are the ones to external validity, that hamper the generality of our results to other scenarios. In our case the only threat in this category is the small scale experimental setup, having compared against two methods over one dataset. However, we have performed small comparisons with other methods and datasets too, and obtained similar results. Moreover, the main goal of our experiments in this paper was to compare different methods, thus the scenario not being so critical to the results.

Finally, the last kind of threats are the construction validity ones, hampering the extrapolation of our results to real-world scenarios. In this case, the range of possible scenarios is potentially infinite, and this threat cannot be fully addressed, but the exploration of how our proposal behaves in other scenarios is matter of future work.


\section{Conclusions}\label{sec:conc}
Knowledge discovery is a key goal when dealing with new data. Using the Unsupervised Cognition algorithm, in this paper we have presented a pipeline for discovering knowledge from it. The practitioner will be able to easily find patterns in the data and refine its dataset using our pattern mining, feature selection and dimensionality reduction proposals. These proposals where presented in this paper and evaluated with multiple experiments. The results of the experiments show how they are state-of-the-art in their respective fields. Thus, this paper presents state-of-the-art techniques for pattern mining, feature selection and dimensionality reduction from an Unsupervised Cognition model.

Specifically, our knowledge discovery pipeline consist in using the presented feature selection technique to select the truly relevant features of the dataset, then use these features to perform dimensionality reduction to train a new Unsupervised Cognition model, and finally use that newly trained Unsupervised Cognition model to extract relevant patterns, with our pattern mining proposal, for the practitioner to explore. Using this pipeline we are able not only to get better, more relevant patterns, but also to build predictive models with a huge increase in accuracy with respect to a predictive model that uses all the data. For example, in our experiments, we were able to build predictive models able to obtain an increase in accuracy of almost $10\%$, raising it from around $80\%$ to almost $90\%$.

As future work, we would like to explore other ways of discovering knowledge using the Unsupervised Cognition algorithm. We would also like to refine our proposals to be more efficient. We would like to explore how our proposal behaves in real-world scenarios too. And we would like to improve Unsupervised Cognition to avoid its dependency on the input order.

\begin{credits}
\subsubsection{\ackname}
We want to thank Daniel Pinyol for his help building the code of our proposals, and Daniel Pinyol and Pere Mayol for our insightful discussions about the topic.

\subsubsection{\discintname}
The authors have no competing interests to declare that are
relevant to the content of this article.
\end{credits}
%
%
%
\bibliographystyle{splncs04}
\bibliography{biblio}

\end{document}